\documentclass[10pt,twocolumn,letterpaper]{article}

\usepackage{cvpr} 
\definecolor{cvprblue}{rgb}{0.21,0.49,0.74}
\usepackage[pagebackref,breaklinks,colorlinks,allcolors=cvprblue]{hyperref}
\usepackage{color}
\usepackage{xcolor}
\usepackage[table]{xcolor}
\usepackage{color,soul}
\usepackage{multicol}
\usepackage{fontawesome5}
\usepackage[accsupp]{axessibility}

\usepackage{multirow}
\newcommand{\pub}[1]{{\color{gray}{\small{[{#1}]}}}}
\usepackage{pifont}
\newcommand{\cmark}{\ding{51}} 
\newcommand{\xmark}{\ding{55}} 

\def\paperID{20692	
} 
\def\confName{CVPR}
\def\confYear{2026}

\title{\underline{B}-MoE: A \underline{B}ody-Part-Aware Mixture-of-Experts “All Parts Matter” Approach to Micro-Action Recognition}

 \author{\parbox{16cm}{\centering
    {\large Nishit Poddar$^{1,3{\color{red}*}}$,  Aglind Reka$^{1,2{\color{red}*}}$, Diana-Laura Borza$^{1}$, Snehashis Majhi$^1$, Michal Balazia$^1$,   Abhijit Das$^3$,  François Brémond$^{1,2}$}\\
    {\normalsize
    $^1$ INRIA \quad
    $^2$ Université Côte d’Azur  \quad
    $^3$ Birla Institute of Technology \& Science, Hyderabad}}\\
    \vspace{0.3cm}
    \small{{\color{red}* These authors contributed equally to this work.}} \quad 
    Code: \href{https://github.com/aglindreka/B-MoE}{\faGithub\ https://github.com/aglindreka/B-MoE}
    \\
    \small{{{Email: \url{aglind.reka@inria.fr, francois.bremond@inria.fr, abhijit.das@hyderabad.bits-pilani.ac.in }}}}
}
\begin{document}
\maketitle
\begin{abstract}

Micro-actions, fleeting and low-amplitude motions, such as glances, nods, or minor posture shifts, carry rich social meaning but remain difficult for current action recognition models to recognize due to their subtlety, short duration, and high inter-class ambiguity. In this paper, we introduce \textbf{B-MoE}, a Body-part-aware Mixture-of-Experts framework designed to explicitly model the structured nature of human motion.
In \textbf{B-MoE}, each expert specializes in a distinct body region (head, body, upper limbs, lower limbs), and is based on the lightweight \textbf{Macro–Micro Motion Encoder} (M3E) that captures long-range contextual structure and fine-grained local motion. A cross-attention routing mechanism learns inter-region relationships and dynamically selects the most informative regions for each micro-action. B-MoE uses a \textbf{dual-stream encoder} that fuses these region-specific semantic cues with global motion features to jointly capture spatially localized cues and temporally subtle variations that characterize micro-actions. Experiments on three challenging benchmarks (MA-52, SocialGesture, and MPII-GroupInteraction) show consistent state-of-the-art gains, with improvements in ambiguous, underrepresented, and low-amplitude classes.

\end{abstract}

\begin{figure*}[t]
\center
\includegraphics[width=1\textwidth]{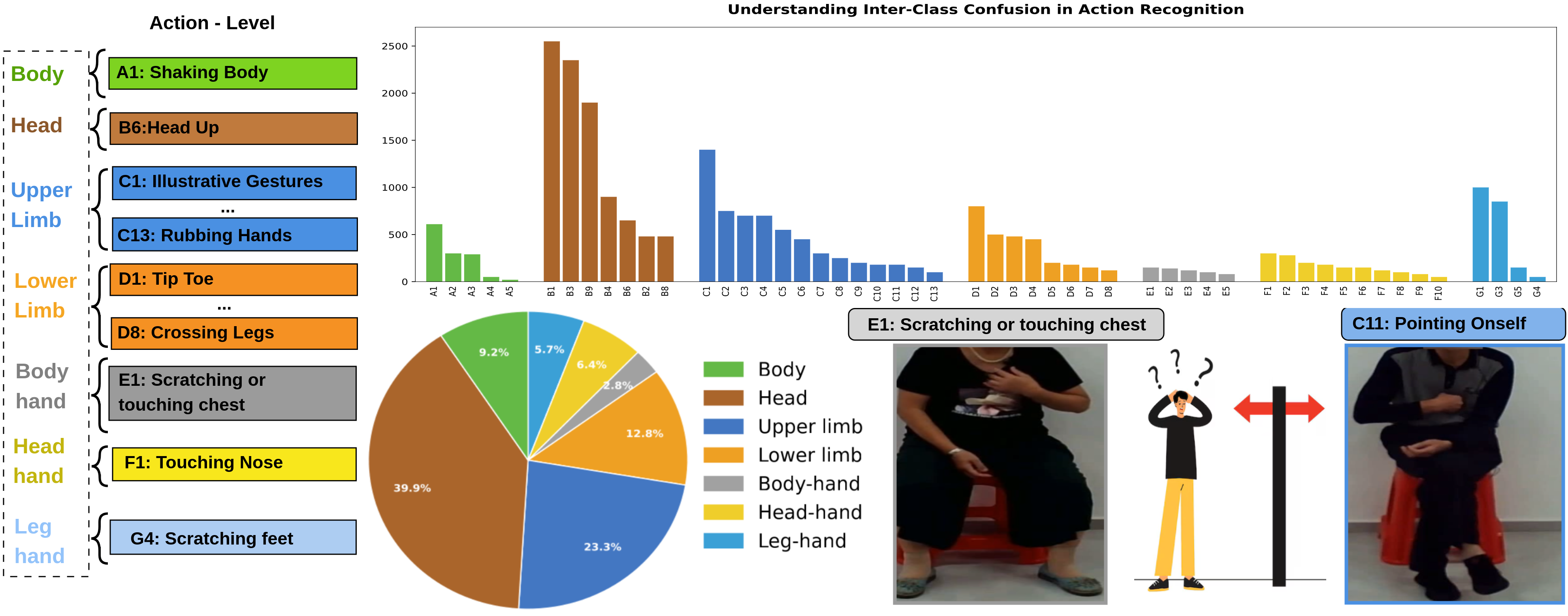}
\caption{Visualization of MA categories illustrating fine-grained motion variations and the challenges of class imbalance and inter-category ambiguity. Example actions (E1: \textit{Scratching or touching chest}, C11: \textit{Pointing oneself}) that appear visually similar but belongs to different semantic categories \cite{guo2024benchmarking}.}
\label{fig1}
\end{figure*}

\section{Introduction}
{We communicate a great deal through body movement without speaking a single word. However, in face-to-face interactions, many of these gestures are nearly imperceptible, such as a head tilt, a shrug or a subtle shift in posture. These \emph{micro-actions} (MA) encode subtle cues of confidence, hesitation, and emotion, and their recognition contributes to a more detailed understanding of human behavior, intent, emotion, and social interactions~\cite{liu2021imigue,chen2023smg,li2024mmad}.

Unlike conventional action recognition, which focuses on clearly distinguishable movements, \emph{micro-action recognition (MAR)} deals with fleeting, fine-grained behaviors that are often subtle, brief, and have overlapping semantics. These behaviors frequently arise from reflexes or specific situational cues, influenced by environmental and social factors. 
A nod or a twitch can be associated with different actions, but the underlying source of motion often originates from different body parts and follows a consistent body-to-action hierarchy. Modeling this articulation helps MAR systems capture coordinated motion across regions such as the head, torso, and limbs. However, most existing approaches treat these actions as flat categories, overlooking both the spatial dependencies between body regions and the localized nature of subtle motion cues~\cite {guo2024benchmarking}. As a result, models that rely on holistic representations often struggle to isolate informative signals from the background or other static regions, and to differentiate between highly similar micro-movements within the same body region.

Another challenge in MAR lies in the imbalance and variability of the data. Datasets such as MA-52~\cite{guo2024benchmarking}, SocialGesture~\cite{cao2025socialgesture}, and MPII-GroupInteraction (MPII-GI)~\cite{balazia2022bodily, mueller18_etra, mueller21_mm} capture a wide spectrum of human movements, from short, dynamic gestures to long, static postures, and introduce variation in both temporal scale and class frequency. This dataset imbalance makes MAR more challenging to capture the rare, yet distinctive motion patterns characteristics of MAR. To address these challenges, we propose \textbf{B-MoE}, a \textbf{body-part-aware MoE} framework for MAR. 

The proposed approach exploits the natural structure of the human body by analyzing motions from localized regions (body, head, upper and lower limbs), and it allows the model to focus on the subtle movements and the discriminative cues within each region. Hence, in contrast to existing works, our model suppresses background interference, strengthens the detection of fine-grained motion cues, and improves discrimination between ambiguous action classes.
In B-MoE, each body part is processed by a dedicated expert built upon our lightweight \textbf{Macro–Micro Motion Encoder (M3E)}, which is designed to capture both long-range contextual structure and fine-grained motion cues. A cross-attention module selects and fuses informative region-wise semantic cues, which are subsequently integrated with global motion features. 
The framework addresses the main MAR challenges - subtlety, ambiguity, and class imbalance - as region-specialized experts amplify fine local cues, the cross-attention mechanism suppresses irrelevant regions, and the dual-stream design provides complementary semantic and motion evidence.

Our approach is built around the following key contributions:
\begin{itemize}

\item We design a \textbf{Body-part-aware Mixture-of-Experts} framework for efficient MAR, in which lightweight experts are dedicated to different body regions (head, body, upper limbs, lower limbs). The proposed framework uses a cross-attention scheme in which the semantic encoder features query each expert, allowing the system to select and weight region outputs according to the motion cues present in the video. This allows the model to focus on the body regions involved in the MAs and downweight irrelevant ones, improving the modeling of subtle and potentially ambiguous region-specific motions.

\item In MAR, discriminative cues may occur as brief subtle motions or as slow variations within the same body region. To address this, we introduce the \textbf{Macro–Micro Motion Encoder (M3E)} as the backbone for all region experts. M3E combines long-range temporal attention with fine-grained local motion reasoning, to allow experts to model both prolonged poses and rapid micro-movements.

\item We conduct extensive experiments on three socially contextual micro-action benchmarks (MA-52, MPII-GI and SocialGesture), and we achieve notable $F1_{macro}$ accuracy gains of \textbf{+4.32}\%, \textbf{+3.35}\%, and \textbf{+1.17}\%, respectively. Our analysis further shows that B-MoE remains robust under class imbalance and substantially improves recognition of \textit{subtle} and \textit{ambiguous} actions.

\end{itemize}

\section{Related Work}

 Understanding micro-actions, the brief, fine-grained movements that reveal social and emotional cues, is an important direction in human-centered video analysis ~\cite{guo2024benchmarking}. Unlike full-body actions, these subtle motions involve coordinated activity across multiple body parts and last just a few frames, and therefore require models that can capture both precise spatial details and rapid temporal changes.

Early work on MAR primarily used general-purpose deep learning architectures, such as 2D/3D CNNs~\cite{tran2015learning, lin2019tsm}, graph convolutional networks~\cite{yan2018spatial}, and more recent transformer-based approaches~\cite{liu2022video}. 
Several studies focused on strengthening local motion cues for MAR: the dual-branch network in~\cite{mi2020dual} introduces a subtle-motion detector to extract mid-level CNN features, while the follow-up model in~\cite{mi2022recognizing} fuses multi-layer local features to improve the extraction of fine body-part movements. MANet~\cite{guo2024benchmarking} augments ResNet~\cite{he2016deep} with Squeeze-and-Excitation (SE) blocks and the Temporal Shift Module (TSM), but SE emphasizes global channel responses over fine local details. Moreover, TSM operates within a fixed temporal window by shifting features across neighboring frames, and this temporal blending can smooth-out rapid micro-movements and reduce the sensitivity to subtle temporal changes.

Although these models have improved micro-action recognition, they still fail to capture the fleeting, fine-grained motion patterns that define micro-actions. This limitation is evident on socially nuanced MAR benchmarks such as MA-52~\cite{guo2024benchmarking}, SocialGesture~\cite{cao2025socialgesture}, and MPII-GI~\cite{balazia2022bodily,mueller18_etra}, where subtle movements and contextual cues are dominant (see ~\cref{fig1}).

Existing approaches have two major limitations: (1) they struggle to model the \textit{rapid micro-scale temporal} dynamics characteristic of MA, and (2) they lack sufficiently \textit{detailed spatial representations} to distinguish highly similar actions. Most models operate on full-frame features and do not explicitly model movements at the level of individual body regions, therefore they are unable to focus on the specific limbs or joints where subtle cues occur. These limitations motivate our region-aware MoE design, which explicitly encodes body-part-specific motion patterns, and our dual-stream formulation, which captures both abrupt temporal changes and fine-grained spatial details.

 \textbf{Mixture of Experts (MoE)} follows a \textit{divide-and-conquer} principle, where a gating network activates only a small subset of specialized experts for each input~\cite{mu2025comprehensive,puigcerver2023sparse,lee2022sparse,fedus2022switch}. This conditional computation enables efficient scaling and expert specialization while maintaining computational efficiency. Furthermore, expert assignments are often difficult to interpret, as gating decisions do not always align with semantically meaningful sub-tasks. Recently, MoE has been explored in action recognition and detection to model heterogeneous motion dynamics and modality-specific cues~\cite{huang2024occluded}. \textbf{GaitMoE}~\cite{huang2024occluded} applies temporal and action experts to handle occluded gait sequences, while \textbf{GS-MoE}~\cite{amicantonio2025mixture} employs class-specific experts guided by a temporal Gaussian splatting loss to enforce temporal consistency under weak supervision. \textbf{MixANT}~\cite{wasim2025mixant} integrates MoE into a Mamba-based sequence model, where expertized state-transition matrices capture observation-dependent dynamics for dense action anticipation. Finally, \textbf{MoNE}~\cite{jain2024mixture} introduces hierarchical routing, processing visual tokens of varying importance through nested experts of different capacities. Although MoNE efficiently routes tokens through nested experts of varying capacities, it assumes that many tokens are redundant and can be processed by cheaper experts. In fine-grained motion tasks, such as micro-action recognition, nearly all tokens are informative, reducing this redundancy. As a result, MoNE’s efficiency advantage diminishes, and the model may struggle to capture subtle, localized motion cues effectively.

In contrast to previous approaches, we propose a \textbf{body-part aware MoE architecture}, in which each expert operates on features extracted from different body parts (head, body, upper limbs, lower limbs), covering the full body structure. The experts have been pretrained only on the classes that belong to that experts and the novelty is that B-MoE during fine-tuning learns how to recognize the interpendencies as the hand touching the leg where only two experts should be activated. The model relies on a semantic encoder stream~\cite{wang2023videomae} to capture rich semantic spatio-temporal representations and motion encoder stream~\cite{guo2024benchmarking} to better encode short, transient motion characteristics of micro-actions. Expert selection and routing are handled by a cross-attention mechanism, which routes the information from different body parts to emphasize the most discriminative cues for each micro-action.

\section{Methodology}

\subsection{Solution overview}

Given a video sequence \( \mathcal{V} = \{ f_1, f_2, \ldots, f_T \} \), where \( f_i \) denotes the \( i^{th} \) frame, the goal of Micro-Action Recognition (MAR) is to predict a micro-action label $\hat{Y_{\text{act}}} \in \mathcal{Y}$. 

\begin{figure}
 \centering
 \includegraphics[width=0.48\textwidth]{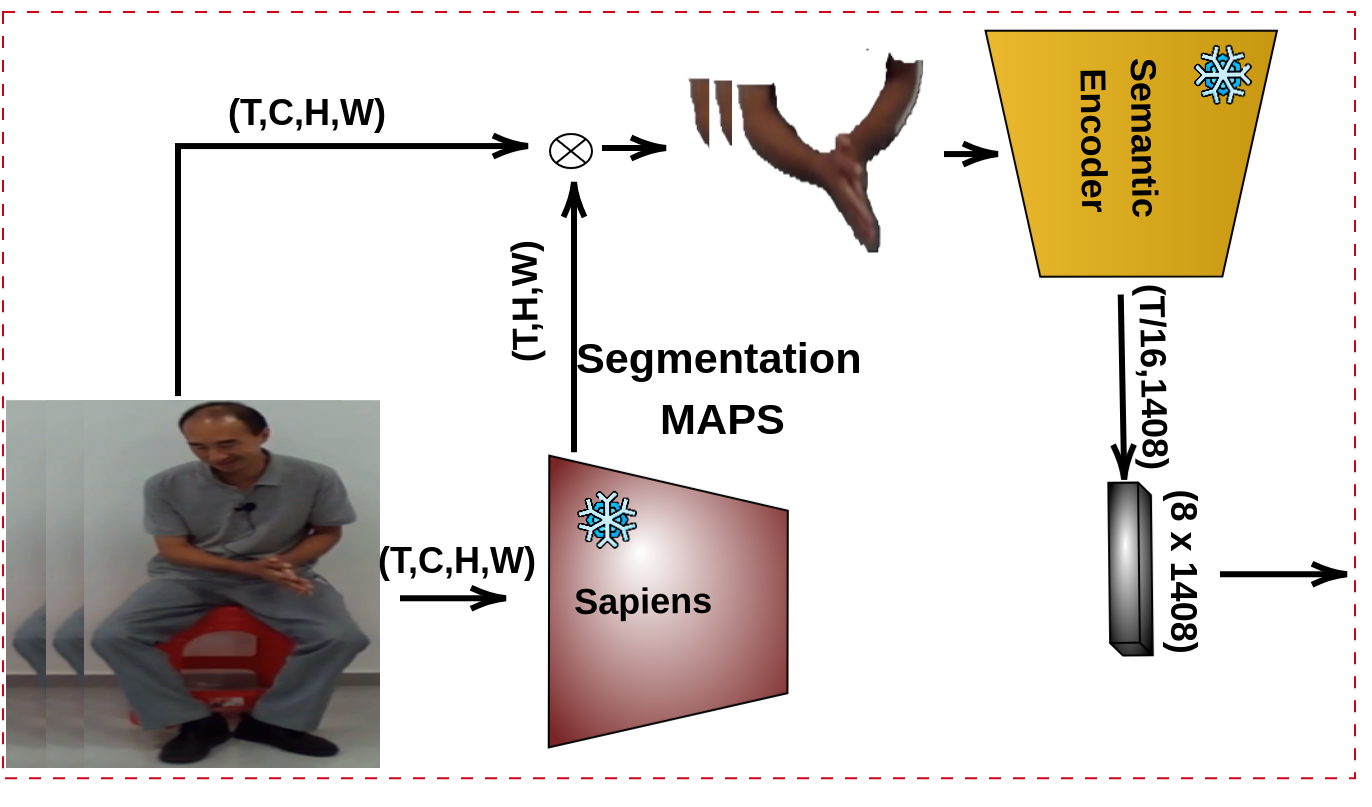}
 \caption{\textbf{Semantic Branch:} 
 Using SAPIENS \cite{khirodkar2024sapiens}, we segment each frame, derive the crop around the target body part (upper limb in this example), and apply the corresponding mask to the cropped region. The resulting cropped and masked video is 
then processed by VideoMAE-V2, pretrained on Kinetics. 
 }
 \label{fig:semantic_branch}
\end{figure}

\begin{figure*}

 \centering
 \includegraphics[width=0.98\textwidth]{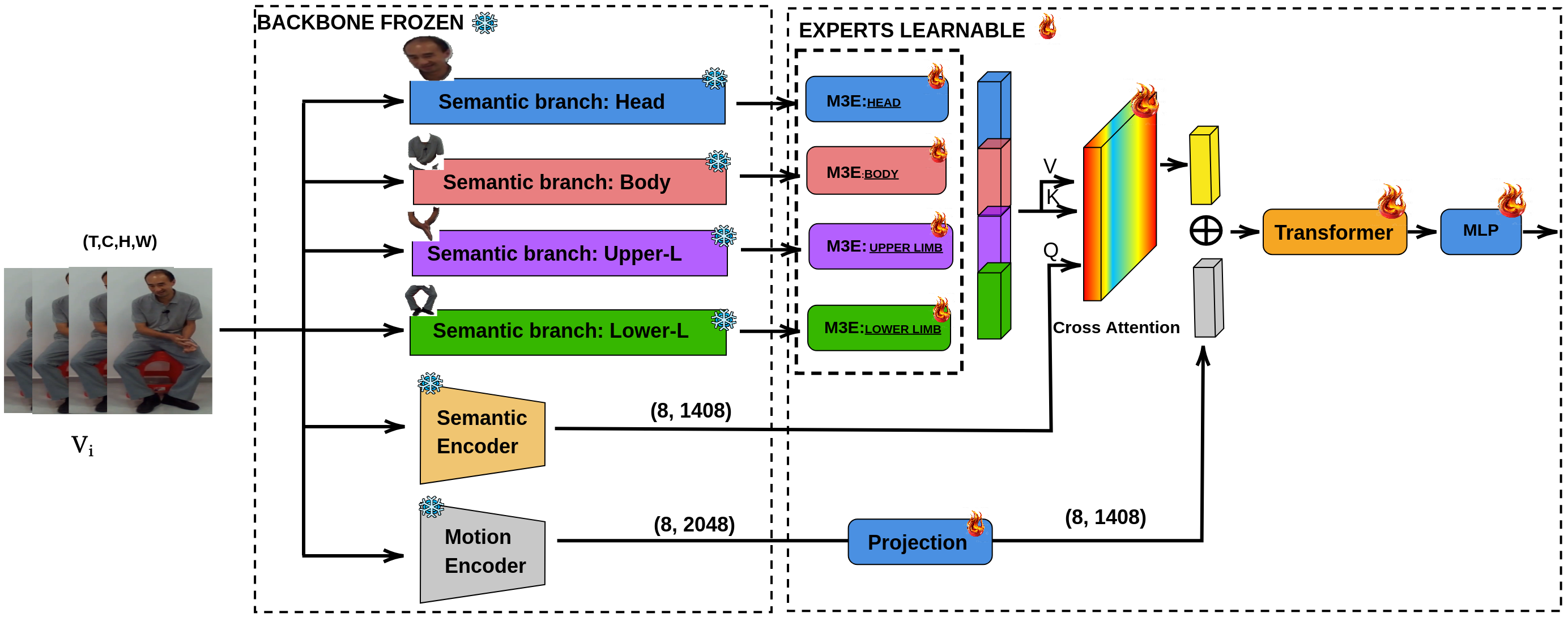}
 \caption{\textbf{B-MoE:} A dual-stream encoder extracts region-conditioned semantic features using semantic encoder and global motion encoder. The semantic stream is routed through a region-aware MoE, where each expert specializes in modeling micro-movements within a specific body region. A cross-attention fusion head integrates expert outputs with motion saliency from the global stream, and a transformer-MLP classifier produces the final predictions. }
 \label{sol_overview}
\end{figure*}



The motion encoder stream operates on full-frame sequences to extract short-term, fine-grained motion patterns through a ResNet~\cite{he2016deep} backbone enhanced with Squeeze-and-Excitation (SE) and the Temporal Shift Module (TSM) modules. It extracts a global representation of the body dynamics and captures how movements evolve over time across the entire frame.

In parallel, the semantic encoder stream relies on VideoMAE-V2~\cite{wang2023videomae} to encode region-specific crops. It encodes the cropped and masked snippets, and outputs a set of region-conditioned semantic embeddings.
This allows the model to focus on the subtle, localized appearance and micro-motion cues that are easily overshadowed in global representations. 
These representations form the input of the MoE architecture, where region-aware mixture of experts further refine the semantics before fusion with motion cues.

\subsection{Region-Aware Mixture of Experts}
\vspace{-3mm}
MAs often originate from specific body regions (e.g., the hands, head, or upper torso), yet manifest differently depending on the surrounding context. To capture this, we introduce a region-aware mixture of experts architecture, as shown in~\cref{sol_overview}, that decomposes the representation learning process along body-regions.
We begin by extracting per-frame human semantic segmentation maps using Sapiens~\cite{khirodkar2024sapiens} (see~\cref{fig:semantic_branch}). These maps are used to localize each target body region (head, body, upper limbs, and lower limbs) by computing the corresponding bounding boxes directly from the segmentation labels. For each identified region, the frame is cropped and masked to retain only the pixels belonging to that specific body part, producing region crops \( R_k \). The semantic embeddings of these cropped regions are passed to the semantic encoder (VideoMAE-V2), and then routed to a set of \textit{lightweight} experts \( \{E_k \mid 1 \le k \le K\} \), where each expert specializes in modeling micro-actions occurring within a specific region \( k \in \{\text{head}, \text{body}, \text{upper limb}, \text{lower limb}\} \) and \( K = 4 \) denotes the total number of experts. This region-specific specialization allows each expert to focus on the subtle motion patterns and appearance cues that characterize micro-actions within its designated region to encourage specialization, while avoiding interference from unrelated body parts. During the final end-to-end training, all experts are optimized jointly and fused through cross-attention, allowing the model to learn cross-region dependencies while retaining region-specific expertise.
Each expert is implemented as an M3E(\cref{fig2}) module pretrained on its corresponding body region classes.


\subsubsection{Macro-Micro Motion Encoder (M3E)}

The expert backbone is based on the Macro-Micro Motion Encoder (M3E), a model designed to capture the wide range of temporal scales present in MA, which may span from brief motions (e.g., \emph{Retracting arms}, \emph{Shaking head}) which are 1.3s in average to prolonged activities (e.g., \emph{Arms akimbo} of 4.3s, \emph{Scratching back} of 3.3s, or even minute-long poses such as \emph{legs crossed}). To handle this duration variability, \textbf{M3E} (~\cref{fig2}) combines long-range contextual modeling with short-term motion sensitivity. 

\begin{figure}[t]
\centering
\vspace{-2.8mm}
\includegraphics[width=0.48\textwidth]{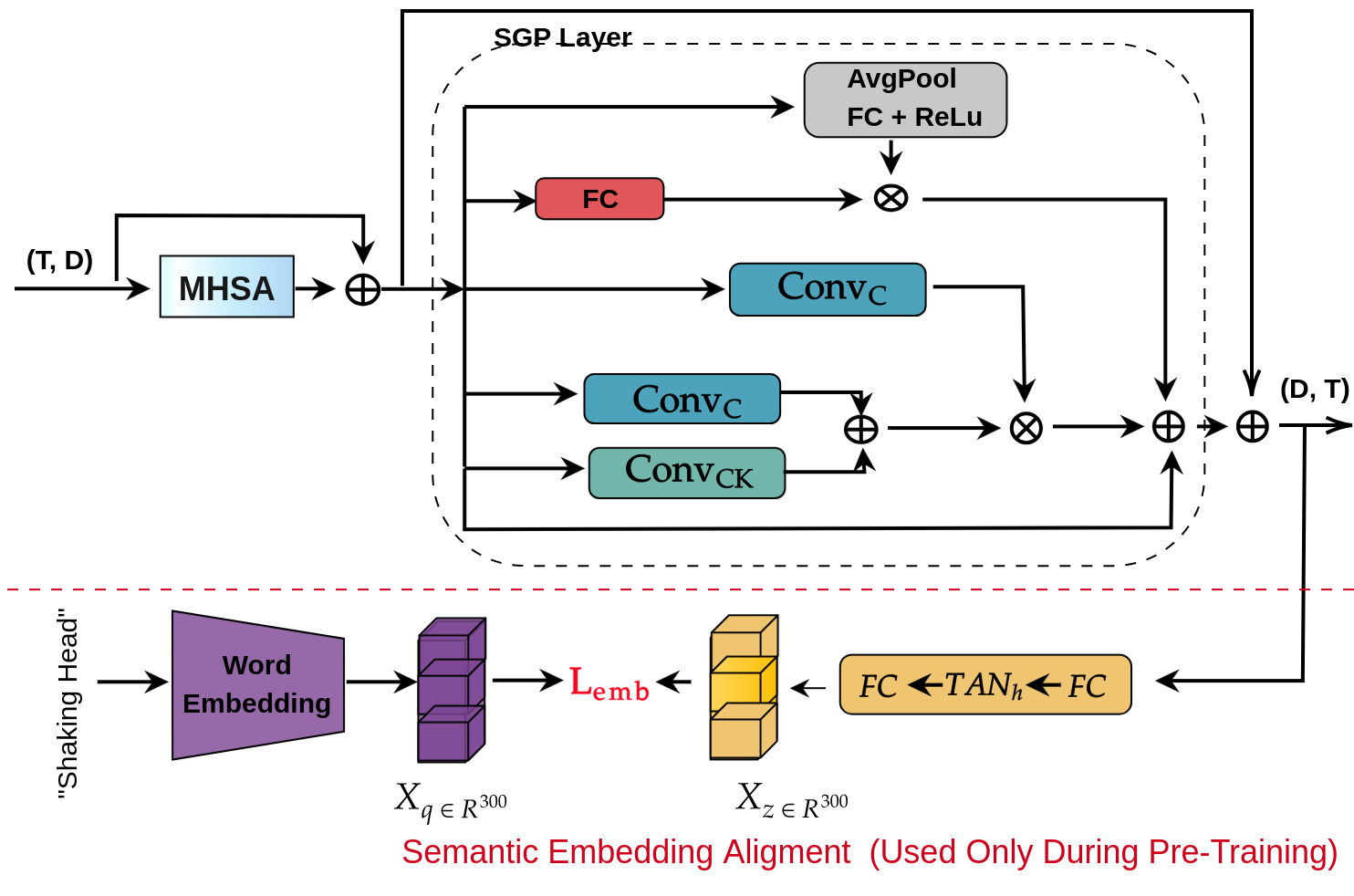}
\vspace{-2.8mm}
\caption{\textbf{Macro-Micro Motion Encoder (M3E).} The input sequence is processed with multi-head self-attention to capture global temporal dependencies, followed by an SGP module~\cite{shi2023tridet} for fine-grained local motion reasoning. During pre-training, a semantic alignment loss ($\mathcal{L}_{\text{emb}}$) aligns learned features with word embeddings of action labels.}
\label{fig2}
\vspace{-2.8mm}
\end{figure}

The module first applies a multi-headed self-attention (MHSA) layer to capture global dependencies across the entire sequence and therefore to allow the model to relate temporally distant but semantically relevant cues. Next, a Scalable-Granularity Perception (SGP) module~\cite{shi2023tridet} refines the representation through two fine-grained operations: frame-level refinement, to capture instantaneous micro-movements, and short-window temporal modeling, to extract local motion patterns over a few frames.
The MHSA is placed before the SGP, to ensure that the SGP refinement operates on features already contextualized over the full sequence, and that the local motion is extracted without losing access to broader intervals.

\subsection{Expert Fusion and Final Prediction}

To aggregate region-aware mixture-of-expert outputs, we employ a cross-attention module that uses the \textit{global embedding} \( z_g \), extracted from the whole body using VideoMAE-V2 as the query, while the keys and values correspond to the region-specific expert representations $\{ z_k\ | 1 \le k \le K\}$. This enables the framework to determine each expert’s relevance conditioned on the full-scene semantics, allowing it to selectively attend to the body regions most informative for the current micro-action. Cross-attention produces a context-aware expert aggregation:
\begin{equation}
 \tilde{z} = CrossAttn(Q=z_{g},K=z_{1:K},V=z_{1:K}),
\end{equation}
where $z_{g}$ is the VideoMAE-V2 feature extracted from the uncropped full-frame snippet, $z_k$ is the $k^{th}$ expert output and $K$ is the number of experts.

The fused semantic representation $\tilde{z}$ is then combined with the motion embedding $z_m$ from MANet via a residual addition to produce a global-local representation that encodes both subtle spatial variations and fine-grained temporal dynamics. Before fusion, the MANet motion embedding is passed through a linear projection layer to match the dimensionality of the semantic embedding.
The fused representation is passed through a lightweight transformer encoder $TE$, to further refine temporal dependencies and allow more interaction between experts and motion cues.

Finally, the transformer output is fed into a multi-layer perceptron (MLP) classifier that predicts the fine-grained micro-action labels $\hat{Y_{\text{act}}}$:
\begin{equation}
 \hat{Y_{\text{act}}} = MLP(TE(\tilde{z} + W_m z_m)),
\end{equation}
where $+$ denotes element-wise addition, and $W_m$ is the linear projection layer that aligns the dimension of $z_m$ with $\tilde{z}$.

\subsection{Training Objective}
The training process is done in two stages. First, we \textbf{pretrain} each expert individually on the subset of classes whose discriminative cues originate primarily from its corresponding body region (e.g. \textit{Nodding} $\rightarrow$ \textit{Head}, \textit{Scraching arm} $\rightarrow$ \textit{Upper limb}) to encourage region-specific motion modeling.
The pretraining objective combines a standard classification loss $\mathcal{L}_{cls}$ on region-specific classes and with an \textbf{embedding alignment loss} ($\mathcal{L}_{emb}$) to align the semantic distance between a video and its corresponding action label. Specifically, each action label is represented using \textbf{GloVe word embeddings}~\cite{pennington2014glove}. 
The embedding vector of an action label, denoted as $\mathbf{X}_q \in \mathbb{R}^{300}$, is obtained by average pooling the embeddings of the words describing the action. 
The averaged visual features of each video are projected into the same embedding space, resulting in $\mathbf{X}_z \in \mathbb{R}^{300}$. 
A semantic constraint is then enforced by minimizing the \textbf{Euclidean distance} between $\mathbf{X}_q$ and $\mathbf{X}_z$:
\begin{equation}
\mathcal{L}_{emb} = \|\mathbf{X}_q - \mathbf{X}_z\|^2.
\end{equation}
\noindent The pretraining objective becomes:
\begin{equation}
\mathcal{L} = \mathcal{L}_{cls} + \alpha \mathcal{L}_{emb},
\end{equation}
where $\alpha$ is a balancing hyperparameter that controls the contribution of the embedding loss. We are using $\alpha$=50 in our experiments. The second step is the \textbf{end-to-end training of the full model}, where we use the standard cross-entropy loss $\mathcal{L}_{cls}$.

\section{Experiments}

\subsection{Datasets}
\paragraph{MA-52 dataset} \cite{guo2024benchmarking} is a large-scale human dataset captured in an interview setting, comprising 22.4K video samples, collected from 205 individuals, with detailed annotations provided at two semantic levels: 7 body-level classes and 52 fine-grained action-level categories. The dataset captures subtle, micro-behaviors across different body regions, such as the head, upper limb, and lower limbs. In our experiments, we use the provided official data splits, which include 11,250 training samples, 5,586 for validation, and 5,586 for testing.

\textbf{MPII-GroupInteraction dataset} \cite{balazia2022bodily, mueller21_mm, mueller18_iui, mueller18_etra} is a rich resource for action recognition in social contexts, and comprises 22 multi-person group discussions (each around 20 minutes) with 78 participants and a total 26 hours of video (2.87M frames). Interactions were recorded using eight synchronized cameras. The following dataset~\cite{balazia2022bodily} includes annotations for 15 fine-grained bodily action classes focused on limb and torso movements, derived from the Ethological Coding System for Interviews (ECSI). In our experiments, we also include four additional classes: crouch, relax, lean towards left, and lean towards right which were included in the dataset annotations but not considered in the original paper’s reported results due to their under representation.
MPII-GI includes 16815 training samples and 6765 validation samples.

\textbf{SocialGesture dataset}\cite{cao2025socialgesture} focuses on hand and arm gestures that arise in multi-person interactions. It focuses on four deictic gestures: pointing, showing, giving, and reaching, which are key for joint attention and object-mediated communication. The dataset contains 293 training videos and 79 test videos, which are further segmented into short clips of 2–5 seconds, yielding 7,944 training clips and 1,945 testing clips. 
Gesture annotations include temporal boundaries, key frames, spatial bounding boxes for both the initiator and target, and corresponding context descriptions. 

\subsection{Implementation Details \& Performance Measure}
For model training, we set the SGD optimizer with a learning rate of 0.00125, a momentum of 0.9, a weight decay of 1e\textsuperscript{-4}, and a batch size of 10 for
model training. The learning rate is reduced by a factor of 10
at the 30\textsuperscript{th} and 60\textsuperscript{th} epochs. The model has been trained for 500 epochs. The pretraining of each expert follows the same configuration as outlined above.
In all our experiments, we use Top-1 accuracy and F1\textsubscript{macro} as our primary evaluation metrics. Top-1 denotes the overall classification accuracy, while F1\textsubscript{macro} computes the average F1 score per class, treating each class equally. We chose F1\textsubscript{macro} to address class imbalance and the long-tail distribution in the data.

\subsection{Comparison with the state-of-the-art}

\begin{table}[t!]
\centering
\tabcolsep 2pt
\resizebox{1.0\linewidth}{!}{
\begin{tabular}{l|c|c}
\hline
Method & Top-1 & F1$_{macro}$\\ \hline
TSN\hspace{0.2em}\pub{ECCV'16} \cite{wang2016temporal} & 34.46 & 28.52\\
TIN\hspace{0.2em}\pub{AAAI'20} \cite{shao2020temporal} & 52.81 & 39.82\\
TSM\hspace{0.2em}\pub{ICCV'19} \cite{lin2019tsm} & 56.75 & 40.19\\
C3D\hspace{0.2em}\pub{ICCV'15} \cite{tran2015learning}
 & 52.22 & 40.86 \\
I3D\hspace{0.2em}\pub{CVPR'17} \cite{carreira2017quo}  & 57.07 & 39.84\\ 
SlowFast\hspace{0.2em}\pub{ICCV'19}  \cite{feichtenhofer2019slowfast} & 59.60 & 44.96\\
VSwinT\hspace{0.2em}\pub{CVPR'22}  \cite{liu2022video} & 57.23 & 38.53\\
TimesFormer\hspace{0.2em}\pub{ICML'21}  \cite{bertasius2021space}
 & 40.67 & 34.38\\ 
Uniformer\hspace{0.2em}\pub{TPAMI'23}  \cite{li2023uniformer} & 58.89 & 48.01\\ 
PoseConv3D(RGB)\hspace{0.2em}\pub{CVPR'22}  \cite{duan2022revisiting}& 56.30 & 35.80\\ 
PCAN(RGB)\hspace{0.2em}\pub{AAAI'25}  \cite{li2025prototypical} & 60.03 & 43.29\\
MANet*\hspace{0.2em}\pub{TCSVT'24}  \cite{guo2024benchmarking} & 60.90 & 48.98 \\

MoNE\hspace{0.2em}\pub{NeurIPS'24} \cite{jain2024mixture} & 20.04 & 15.08\\
GS-MoE\hspace{0.2em}\pub{ICCV'25} \cite{amicantonio2025mixture} & 56.93 & 42.08\\
\hline
\rowcolor{gray!20}\textbf{B-MoE (ours)} & \textbf{64.54} \textcolor{red}{(+3.64)}& \textbf{53.30} \textcolor{red}{(+4.32)}\\
\hline
\end{tabular}
}
\caption{Performance comparison of MAR on the MA-52 dataset.} 
\
\textsuperscript{*}\footnotesize~Model run on Tesla V100 GPU.
\label{tab:results_ma52}
\end{table}

As shown in~\cref{tab:results_ma52}, MANet \cite{guo2024benchmarking} achieves a Top-1 accuracy of 60.90\%, making it one of the strongest performing baselines. Our model, B-MoE, improves upon this by achieving 64.54\% in Top-1 accuracy, a gain of \textbf{3.64}\% and \textbf{4.32}\% in F1\textsubscript{macro}. This improvement can be attributed to the structured modeling in B-MoE, which relies on expert modules to better capture distinctions between action categories, particularly those involving similar body regions. 

On the MPII-GI dataset (\cref{tab:merged_mpii_socialgesture}), the proposed architecture achieves a +2.57\% improvement in Top-1 accuracy and a +3.35 increase in F1\textsubscript{macro}. Similarly, on the SocialGesture dataset (\cref{tab:merged_mpii_socialgesture}) shows a +1.17 gain in F1\textsubscript{macro} on a highly imbalanced dataset, indicating that B-MoE handles class imbalance more effectively than other models.

To showcase the effectiveness of the proposed B-MoE we also evaluated two other recent MoE models, GS-MoE \cite{amicantonio2025mixture} and MoNE \cite{jain2024mixture}. Specifically, we adapted GS-MoE \cite{amicantonio2025mixture} to our supervised setting, where all labels are available. The results of these models can be observed across MA-52 dataset, and MPII-GI and SocialGesture datasets in~\cref{tab:results_ma52} and~\cref{tab:merged_mpii_socialgesture}, respectively. As observed, in the case of GS-MoE, training all experts on the full body or entire video does not encourage them to focus on specific regions. Since our task involves micro-movements, pretraining each expert independently on particular body parts allows the model to attend more carefully to fine-grained motion in each region.

MoNE allocates compute adaptively across visual-tokens, but without any notion of body structure or localized motion, many subtle motions in MAs will be routed to low-capacity experts. Additionally, MoNE’s expert structure was designed for general video and image datasets, where experts focus on coarse visual features. As shown in~\cref{tab:merged_mpii_socialgesture}, it performs reasonably well on datasets that do not require fine-grained micro-actions, such as SocialGesture, but its accuracy is still lower than our model.

\paragraph{Computational cost.}
In terms of computational efficiency, we compare B-MoE with another MoE architecture, \textbf{GS-MoE}. Our \textbf{B-MoE} requires only \textbf{567.77M} GFLOPs and 
~\textbf{59M} parameters, whereas \textbf{GS-MoE} operates at \textbf{801.26M} GFLOPs with 
~\textbf{80M} parameters. Despite being significantly lighter, B-MoE achieves a \textbf{+7.61\%} Top-1 improvement on MA-52.

\begin{table}[htbp!]
\scriptsize
\centering
{
\begin{tabular}{l|c|c|c|c}
\hline
\multirow{2}{*}{Method} &
\multicolumn{2}{c}{MPII-GI} &
\multicolumn{2}{|c}{SocialGesture} \\
\cline{2-5}
 & Top-1 & F1$_{macro}$ & Top-1 & F1$_{macro}$ \\
\hline
I3D\hspace{0.2em}\pub{CVPR'17}        & 62.24 & 34.62 & 83.54 & 25.54 \\
VSwinT\hspace{0.2em}\pub{CVPR'22}     & 57.16 & 30.73 & 85.45 & 42.58 \\
TimesFormer\hspace{0.2em}\pub{ICML'21}& 49.87 & 17.79 & 84.88 & 45.38 \\
MANet\hspace{0.2em}\pub{TCSVT'24}     & 66.86 & 41.63 & 85.74 & 47.76 \\
MoNE\hspace{0.2em}\pub{NeurIPS'24}    & 30.85 & 4.16  & 82.34 & 22.63 \\
GS-MoE\hspace{0.2em}\pub{ICCV'25}     & 61.73 & 35.11 & 85.81  & 49.35   \\

\hline
\textbf{B-MoE (ours)} &
\shortstack{
  \textbf{69.43} \\
  \textcolor{red}{\scriptsize{(+2.57)}}
} &
\shortstack{
  \textbf{44.98} \\
  \textcolor{red}{\scriptsize{(+3.35)}}
} &
\shortstack{
  \textbf{86.12} \\
  \textcolor{red}{\scriptsize{(+0.31)}}
} &
\shortstack{
  \textbf{50.52} \\
  \textcolor{red}{\scriptsize{(+1.17)}}
} \\
\hline
\end{tabular}
}
\caption{Comparison of MAR performance on MPII-GI and SocialGesture.}
\label{tab:merged_mpii_socialgesture}
\end{table}

\section{Ablation studies}

To systematically evaluate the contribution of each component in B-MoE, we conduct a series of ablation studies on the MA-52 dataset. We first analyze the dual-encoder design to assess the importance of complementary representations. Next, we examine the effect of removing experts to understand their individual contributions, and analyze the cross-attention mechanism through visualization of attention weights to verify expert alignment. Finally, we evaluate B-MoE’s behavior across underrepresented, subtle, and ambiguous actions. The following sections walk through each of these analyzes in detail.

\subsection{Effectiveness of the Dual Stream Encoder}

To evaluate the contribution of the dual-stream encoder, we perform an ablation study that isolates the effect of each encoder, i.e., the semantic encoder and the motion encoder. The results in~\cref{tab:ablation_dual_encoder} allow us to measure how much each stream contributes independently, as well as the performance gain obtained from their interaction. When using only the semantic stream, the model’s performance declines by 5.23\% in Top-1 accuracy and 5.48\% in $F1_{\text{macro}}$ compared to the full dual-stream model, showing that appearance-only cues fail to capture the rapid, fine-grained motions of micro-actions. Relying solely on the motion stream performs slightly better, but still falls short of the full model by 3.64\% Top-1 and 4.32\% $F1_{macro}$. While dynamic cues are important, they lack the semantic context needed to separate visually similar actions. Combining both encoders yields the highest performance, achieving 64.54\% Top-1 and 53.30\% $F1_{macro}$, demonstrating the complementary nature of semantic and motion features.

\begin{table}[htbp]
 \centering
 \begin{tabular}{c|c|c|c}
 \hline
 Semantic & Motion & Top-1 & F1$_{macro}$\\
 \hline
 \cmark & \xmark & 59.31 &  47.82\\
 \xmark & \cmark & 60.90 & 48.98\\
 \hline
 \rowcolor{gray!20}
 \cmark & \cmark & \textbf{64.54}  & \textbf{53.30}\\
 \hline
 \end{tabular}
 \caption{Ablation study on the impact of the dual-stream encoder (semantic and motion) on MA-52 dataset. A \cmark\ indicates that the corresponding stream is included in the model, while an \xmark\ indicates that it is removed.}
 \label{tab:ablation_dual_encoder}
\end{table}

\begin{table}[htbp]
\scriptsize
\centering
\begin{tabular}{c|c|c|c|c|c|c}
\hline
{\#} & Head & Body & Upper-Limb & Lower-Limb & Top-1 & F1$_{macro}$\\
\hline
1 & \xmark & \cmark & \cmark & \cmark &  62.01 &   51.72 \\
2 & \cmark & \xmark & \cmark & \cmark &  64.21 & 52.14  \\
3 & \cmark & \cmark & \xmark & \cmark & 63.76 & 51.94 \\
4 & \cmark & \cmark & \cmark & \xmark &  63.94 &  52.62 \\
\hline
\rowcolor{gray!20}
5 & \cmark & \cmark & \cmark & \cmark &  \textbf{64.54} &  \textbf{53.30} \\
\hline
\end{tabular}
\caption{Ablation experiments on the impact of the experts on MA-52. A \cmark\ indicates that the corresponding expert is included in the architecture, while \xmark \ denotes its removal.}
\label{tab:ablation_experts}
\end{table}

\subsection{Ablation on Region-Aware Experts}

To evaluate the contribution of each expert module, we conduct an ablation study, where we remove one expert at a time (~\cref{tab:ablation_experts}). As in the other experiments, we report Top-1 and $F1_{macro}$ scores. The results show that removing \textit{any} expert consistently reduces performance. We observe that the largest performance drops occur when removing the \textit{Head} expert, with a Top-1 decrease of 2.53\%. This highlights the important role of the head movements and facial cues in micro-action understanding. On the other hand, the \textit{Body} expert shows the lowest performance degradation when removed, indicating that torso-level cues are less distinctive in isolation and often rely on supporting context from the limbs or head. This is aligned with the fact that MAs are usually expressed through finer articulations of the head, arms, or legs rather than large-scale torso motions.
 
 This degradation in performance when experts are removed demonstrates that no single expert dominates the prediction; rather, MAs occur from coordinated whole-body patterns. As expected, the full model (last row) achieves the best performance, which shows that all four experts are necessary to capture the diverse MA patterns present in real-world scenarios.

\cref{fig:cross_attention_weights} visualizes the average cross-attention weights assigned to each expert across all classes in the MA-52 dataset. Each row corresponds to an MA category and each column represents one of the four experts (Head, Upper Limb, Body, Lower Limb). Brighter colors indicate higher attention, which, in turn, reflects greater expert importance for the corresponding action. The heatmap shows a clear region-action correspondence: actions driven by head motion consistently activate the \textit{Head} expert, lower body actions give higher weight to \textit{Lower Limb} expert.

As shown in~\cref{fig:cross_attention_weights}, the first expert (\textit{Head}), highlighted by the red box, primarily focuses on the classes \textit{Bowing head} and \textit{Head-up}, and is activated less for other classes that belong to different expert categories. A similar pattern is observed for the \textit{Upper-Limb} and \textit{Lower-Limb} experts: classes indexed from 11 to 23 correspond to upper-limb actions, whereas classes 24 to 31 correspond to lower-limb actions, both exhibiting stronger activations for their respective regions. In contrast, the classes indexed from 32 to 35 (outlined in a white box) show shared activation between the upper-limb and body experts. These classes, such as \textit{Scratching or touching chest} and \textit{Arms akimbo}, naturally involve coordinated movements between the upper-limb and body, illustrating meaningful cross-expert correlations. An observable pattern in the heatmap is that the \textit{Body} expert is activated less frequently than the others. This reflects the nature of the MA-52 dataset, where many actions involving the body are coupled with movements of the head or limbs, and this indicates that the \textit{Body} expert often provides complementary cues rather than acting as the primary region.


These results validate the core idea of the proposed architecture: different body regions contribute differently depending on the action, and a region-aware MoE architecture captures this structure. Also, they show the specialization and efficiency of the expert routing mechanism: experts are not uniformly engaged, but instead activated only when the underlying motion semantics require their contribution.

\begin{figure}[htbp]
\centering
\includegraphics[width=0.450\textwidth]{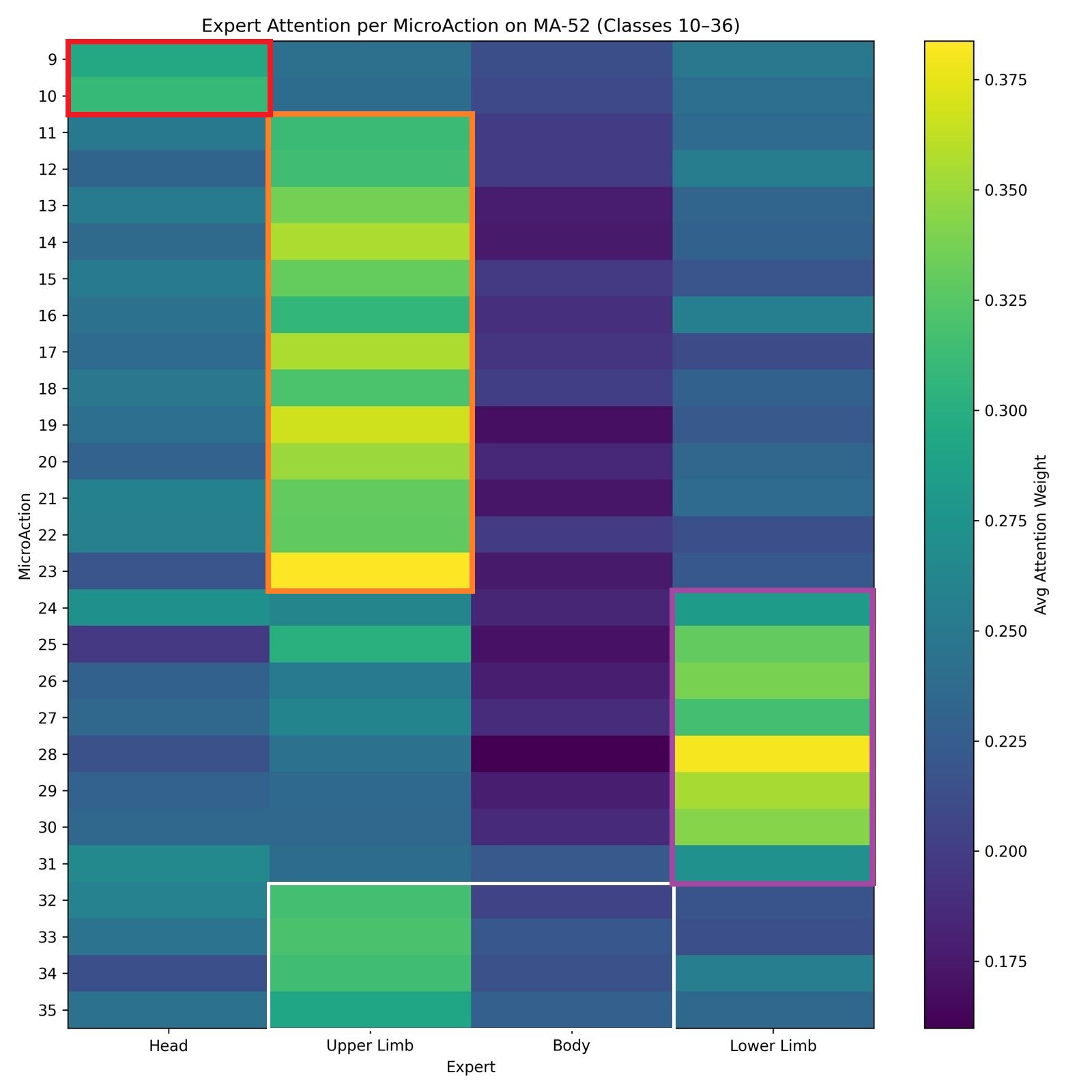}
\caption{Cross-attention heatmap showing expert importance per micro-action class on MA-52. Brighter colors indicate higher expert activation.}
\label{fig:cross_attention_weights}
\end{figure}

\subsection{Ablation Analysis on Ambiguous, Underrepresented, and Subtle Action Classes}

To better understand B-MoE’s performance across ambiguous, underrepresented, and subtle action classes, \cref{fig:Model_comparison} presents a per-class comparison against MANet, the strongest baseline focusing on classes highlighted with different colored boxes. 

\textbf{Black Box (Ambiguous Classes)}: Actions such as \textit{covering mouth} and \textit{covering face} are inherently ambiguous and prone to mislabeling due to overlapping visual cues. For these classes, B-MoE achieves accuracies of 66.67\% and 50.00\%, respectively, whereas MANet predicts no correct instances for these classes (0.00\% for both).

\textbf{Blue Boxes (Underrepresented Classes):} Actions such as \textit{Scratching feet} and \textit{Shrugging}, each account for only 0.18\% of the training set, meaning they appear only a handful of times during learning. Such low occurrence makes it difficult for models to form stable, discriminative representations for these classes. Despite their limited representation, B-MoE achieves accuracies of 60.00\% and 22.00\% on these classes, respectively, compared to MANet’s 30.00\% and 0.00\%. This shows that B-MoE retains generalization capability even for rare and underrepresented action classes.

\textbf{Red Box (Subtle Classes)}:
Classes such as \textit{Nodding} and \textit{Sitting straightly} involve subtle, low-amplitude motions that provide only faint visual cues for recognition. Even in these challenging cases, B-MoE demonstrates a slight performance improvement over MANet. For instance, \textit{Nodding} achieves a meaningful accuracy gain of \textbf{4.68\%}, indicating that B-MoE is more sensitive to small-scale motion variations and can better capture the fine-grained cues characteristic of micro-actions.

\begin{figure}[htbp]
\centering
\includegraphics[width=0.50\textwidth]{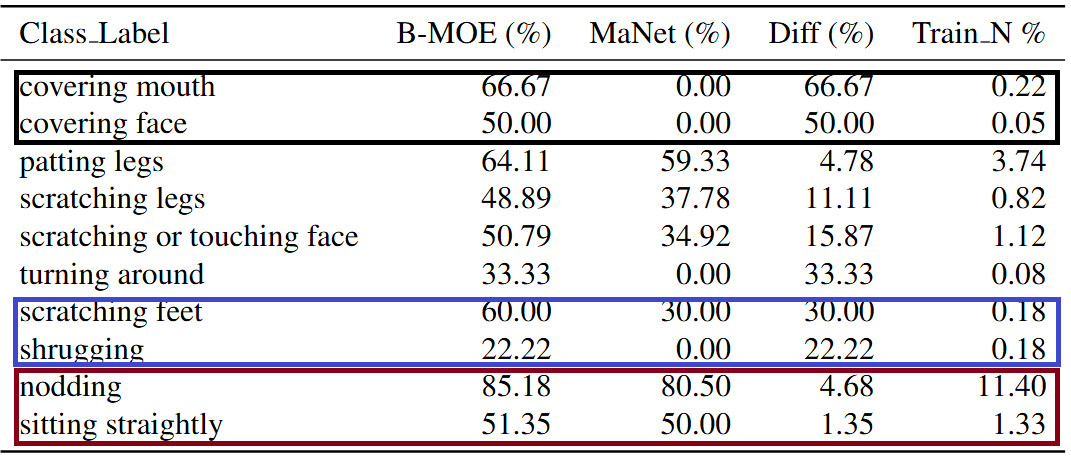}
\caption{Per-class comparison of B-MoE and MaNet on the validation set for MA-52. Classes enclosed by black, blue, and red boxes correspond to \textit{ambiguous}, \textit{underrepresented}, and \textit{subtle} actions, respectively. The column Train\_N indicates the percentage of samples for each class.}
\label{fig:Model_comparison}
\end{figure}

\section{Conclusion}
\vspace{-2mm}
In this work, we introduced a MoE framework for micro-action recognition that structures recognition along body-region boundaries. Each expert is built upon the proposed M3E backbone, which models both global contextual cues as well as fine-grained local motions. The framework relies on a dual-stream encoder that disentangles motion and semantic information: motion encoder extracts full-frame motion features, while semantic encoder provides semantically rich, region-conditioned representations. A cross-attention fusion module integrates these complementary signals in a context-aware manner, allowing the system to highlight the most informative regions for each MA. Extensive evaluation on the three socially contextual MAR benchmarks shows that our approach consistently outperforms other methods, even under class imbalance and subtle inter-class variation.

While our method achieves strong performance, it still relies on manually defined expert configurations. A direction for future work is to enable the model to automatically determine not only which experts to activate, but also how many are needed for a given input. Additionally, extending the framework beyond RGB to incorporate depth, skeleton, or audio modalities could further improve robustness and help disambiguate visually similar micro-actions. 
\textbf{Acknowledgements:} This work was carried out as part of the Intelligent Mapping project, a component of the IRIMA Platforms project, funded by the French ANR (grant no. ANR-22-EXIR-0008). The work also received support from the EUR SPECTRUM of Université Côte d’Azur, France.

{
 \small
 \bibliographystyle{ieeenat_fullname}
 \bibliography{main}
}


\end{document}